\documentclass{article}

\usepackage{microtype}
\usepackage{graphicx}
\usepackage{subfigure}
\usepackage{booktabs} 

\usepackage{bm}

\usepackage[accepted]{icml2025}

\usepackage{amsmath}
\usepackage{amssymb}
\usepackage{mathtools}
\usepackage{amsthm}
\usepackage{float}
\usepackage[textsize=tiny]{todonotes}
\usepackage{hyperref}
\usepackage[capitalize,noabbrev]{cleveref}

\theoremstyle{plain}

\theoremstyle{definition}

\theoremstyle{remark}

\icmltitlerunning{Next-Token Prediction Should be Ambiguity-Sensitive}

\begin{document}

\twocolumn[
\icmltitle{Next-Token Prediction Should be Ambiguity-Sensitive: \\A Meta-Learning Perspective}



\icmlsetsymbol{equal}{*}

\begin{icmlauthorlist}
\icmlauthor{Leo Gagnon}{mila}
\icmlauthor{Eric Elmoznino}{mila}
\icmlauthor{Sarthak Mittal}{mila}
\icmlauthor{Tom Marty}{mila}
\icmlauthor{Tejas Kasetty}{mila}\\
\icmlauthor{Dhanya Sridhar}{mila}
\icmlauthor{Guillaume Lajoie}{mila}
\end{icmlauthorlist}

\icmlaffiliation{mila}{Mila - Quebec's AI Institute. Université de Montréal}

\icmlcorrespondingauthor{Leo Gagnon}{leogagnon@gmail.com}

\icmlkeywords{Machine Learning, ICML}

\vskip 0.3in
]



\printAffiliationsAndNotice{}  
\begin{abstract}
\looseness=-1
The rapid adaptation ability of auto-regressive foundation models is often attributed to the diversity of their pre-training data. This is because, from a Bayesian standpoint, minimizing prediction error in such settings requires integrating over all plausible latent hypotheses consistent with observations. While this behavior is desirable in principle, it often proves too ambitious in practice: under high ambiguity, the number of plausible latent alternatives makes Bayes-optimal prediction computationally intractable. Cognitive science has long recognized this limitation, suggesting that under such conditions, heuristics or information-seeking strategies are preferable to exhaustive inference. Translating this insight to next-token prediction, we hypothesize that low- and high-ambiguity predictions pose different computational demands, making ambiguity-agnostic next-token prediction a detrimental inductive bias. To test this, we introduce MetaHMM, a synthetic sequence meta-learning benchmark with rich compositional structure and a tractable Bayesian oracle. We show that Transformers indeed struggle with high-ambiguity predictions across model sizes. Motivated by cognitive theories, we propose a method to convert pre-trained models into Monte Carlo predictors that decouple task inference from token prediction. Preliminary results show substantial gains in ambiguous contexts through improved capacity allocation and test-time scalable inference—though challenges remain. Code is available \href{https://github.com/leogagnon/metahmm}{here}.
\end{abstract}

\vspace{-7mm}
\section*{Introduction}
\label{sec:intro}
\looseness=-1
A leading explanation for the surprising generalization capabilities of transformer-based \citep{vaswani2017attention} foundation models \citep{bommasani2021opportunities} is that their pretraining distribution resembles a sequence meta-learning problem \citep{brown2020language, xie2021explanation, chan2022data, wang2023large, hahn2023theoryemergentincontextlearning}. In this view, each document in the corpus is governed by latent factors (e.g., topic, world state), and models learn to perform implicit Bayesian inference over these factors to predict tokens effectively across domains.

\looseness=-1
In the idealized setting \citep{ortega2019meta}, each sequence is generated by sampling a task $\theta \sim p^*(\theta)$ and then drawing observations from $p^*(x_{1:T}\mid \theta)$. The next-token predictor that minimizes prediction error in that setting is called the \textit{Bayes-optimal} posterior predictive :
\begin{align}\label{eq:oracle}
p^*(x_t\mid x_{<t})=\int_\theta \underbrace{p^*(x_t\mid x_{<t},\theta)}_{\text{Prediction}} \underbrace{p^*(\theta\mid x_{<t})}_{\text{Inference}}
\end{align}\par
\vspace{-2mm}
\looseness=-1
Thus, training a model to minimize next-token prediction loss (LHS) is encouraged to implicitly perform task inference (RHS)—notably explaining how foundation models can adapt to new tasks at inference time purely by conditioning on a few input examples \citep[In-Context Learning, ICL;][]{brown2020language, panwar2024incontextlearningbayesianprism}.
\begin{figure*}[]
  \centering
    \includegraphics[width=\linewidth]{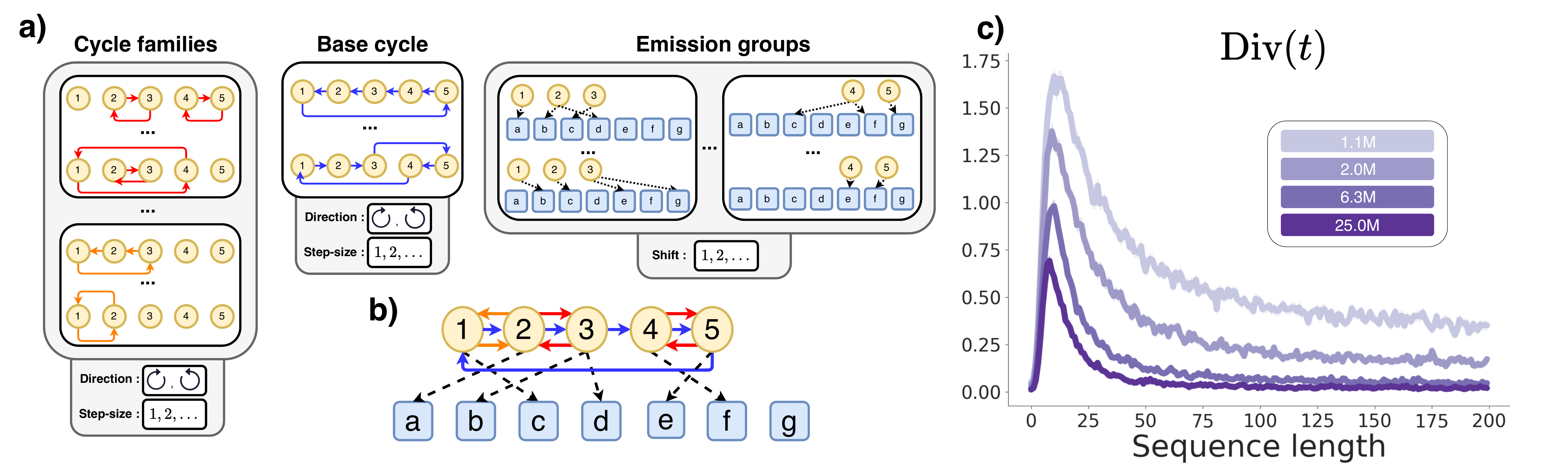} 
    \vspace{-8mm}
    \caption{\textbf{The MetaHMM benchmark.} \textbf{a)} Latent structure of a MetaHMM environment. White rectangles contain mutually exclusive discrete choices $\theta_i$ which together define an HMM $\bm{\theta}$. Yellow circles represent hidden states and blue rectangles represent observable symbols. \textbf{b)} Example of an HMM sampled from the given building blocks. \textbf{c)} $\text{Div}_x(t)$ for different model sizes.}
    \vspace{-3mm}
    \label{fig:metahmm}
\end{figure*}
\looseness=-1
This meta-learning view, however, exposes a fundamental challenge: in high-ambiguity contexts, where the posterior over tasks $p^*(\theta\mid x_{<t})$ has high entropy, prediction becomes inherently harder due to the need to consider many plausible hypotheses $\theta$. In fact, cognitive science has long recognized that Bayes-optimal prediction becomes intractable under resource constraints \citep{lieder2020resource}. Humans respond with ambiguity-aware strategies such as heuristics \citep{binz2022heuristics}, approximate inference \citep{sanborn2010rational}, or information-seeking \citep{friston2017active}. 

\looseness=-1
Following from this, we hypothesize that sequence models which allocate fixed computation per token suffer from poor capacity allocation : the more difficult high-ambiguity predictions receive too little while low-ambiguity ones receive too much. From a statistical learning angle, we suggest that ambiguity-sensitivity may serve as an effective inductive bias for general next-token prediction. 
While it is well understood that prediction under ambiguity causes issues \textit{at inference} in foundation models \citep{liu2023we, keluskar2024llms}, we are the first to ground this problem in meta-learning theory, make links to resource-rational analysis \citep{lieder2020resource} and make the case for poor resource allocation \textit{at training}.

\looseness=-1
To test this hypothesis, we introduce MetaHMM, the first synthetic benchmark for sequence meta-learning with rich compositional latent structure and an exactly computable Bayesian oracle. By evaluating learned next-token predictors against the Bayes-optimal model in \cref{eq:oracle}, we show that Transformers consistently underperform in high-ambiguity settings. Importantly, this gap persists across model sizes, suggesting that scale alone does not resolve ambiguity-related failures. We release our codebase for procedural and scalable MetaHMM generation.

\looseness=-1
To mitigate this issue, we propose a modular predictor that approximates \cref{eq:oracle} using a Monte Carlo (MC) estimator, bootstrapped from a classical autoregressive model. Inspired by human approximate inference \citep{sanborn2010rational}, our method separates task inference from token prediction, introducing useful inductive biases and allowing test-time scaling \citep{snell2024scaling} by increasing the number of MC samples. In doing so, we describe a principled type of test-time scaling as adaptation to posterior entropy—grounded in both Bayesian theory and resource-rational cognition.

\looseness=-1
On MetaHMM, our method consistently outperforms the underlying sequence model in high-ambiguity settings, with gains increasing as more samples are drawn. However, performance gains diminish with larger models, suggesting this approach is especially useful when base models underfit. Adapting our method to naturalistic settings is an important direction for future work. More broadly, exploring additional types ambiguity-aware strategies may be essential for improving robustness and efficiency of foundation models. This work takes a first step in that direction.
\section{The MetaHMM environment}
\label{sec:metahmm}
\looseness=-1
We choose a synthetic environment to isolate and analyze the ambiguity problem without the confounding complexity of natural language. While language corpora resemble a meta-learning distribution, they go far beyond that formalism in ways that would obscure the underlying mechanisms we aim to study. Moreover, no existing sequence meta-learning benchmark possesses the following two properties (a) a non-trivial, structured space of generators that supports meaningful latent inference, and (b) a fully tractable Bayes-optimal predictor (see \cref{sec:related} for further discussion).

\looseness=-1
A MetaHMM environment consists of a family of Hidden Markov Models \citep[HMMs;][]{rabiner1986introduction} where each member is described by a latent code $\bm{\theta}$ which specifies how to build the HMM from a pool of shared building blocks. Concretely, each coordinate $\theta_i$ of $\bm{\theta}$ corresponds to a discrete choice which defines the HMM. Further, one can control the size of a MetaHMM by adding/removing choices; explicit size computation given in \cref{sec:details}. 

The transition matrix of each HMM is composed of one \textit{base cycle} which goes through all hidden states and multiple groups of smaller cycles from \textit{cycle families}. For both the base cycle and cycle families, the direction and speed at which cycles are traversed can change (through a \textbf{Direction} and \textbf{Step-size} variable). Each outgoing edges of a node (after adding all cycles together) have equal probability. The emission matrix of each HMM is built from multiple different \textit{emission groups} together partitioning the hidden states. Additionally, all groups' emission mappings can be cyclically shifted through the \textbf{Shift} variable. See \cref{fig:metahmm}\textbf{a)}.

Importantly, our setup enables efficient and exact computation of the posterior predictive in \cref{eq:oracle} using JAX \citep{bradbury2018jax} implementations of the forward algorithm \citep{linderman2025dynamax}. 

Due to the Markovian nature of HMMs, the ambiguity of $p^*(\theta\mid x_{<t})$ decreases monotonically with sequence length. As a result, the beginning of each sequence corresponds to the high-ambiguity regime—the region where we expect models to perform most poorly. At long context length, when $\theta$ is unambiguous, we expect the Transformer to easily simulate the HMM \citep{rizvi2024simulating}.
\begin{figure*}
  \centering
    \includegraphics[width=\linewidth]{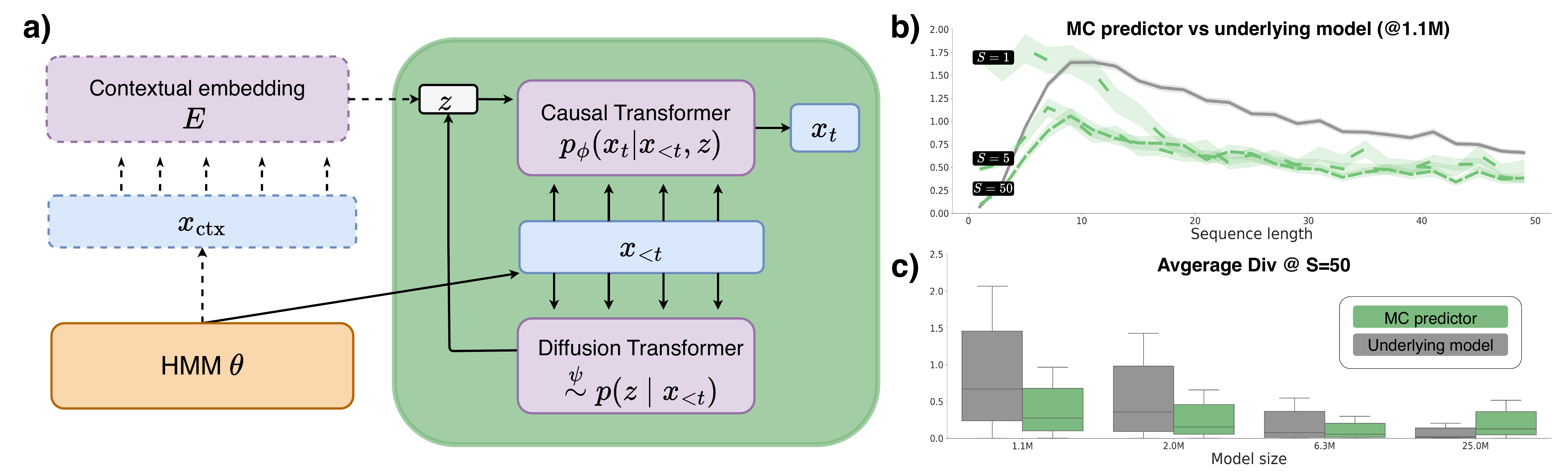} 
    \vspace{-5mm}
    \caption{\textbf{The Monte-Carlo Predictor}. \textbf{a)} Computational and training structure of the MC predictor. The parts in dotted lines are only used to train $p_\phi(x_t\mid x_{<t},z)$, while the content of the green rectangle corresponds to the MC predictor. \textbf{b)} Performance of the MC predictor (green) with various number of samples compared with the original sequence model (gray, 1.1M). Y-axis is $\text{Div}$. \textbf{c)} Average $\text{Div}(t)$ (across sequence length) of the MC predictor (green) with $S=50$ compared to the underlying model (gray) for the different model sizes in \cref{fig:metahmm}.} 
    \label{fig:mc_pred}
    \vspace{-5mm}
\end{figure*}
\subsection*{Evaluation of Transformers}
\looseness=-1
We generate MetaHMM environments of size $\sim$ 12,000 and train 4 different sizes of causal Transformer models $p_\phi$ on them (\cref{sec:details} for more details). We train on sequences of length $T=200$ and evaluate each model by computing a symmetrized KL divergence between its posterior predictive distribution and that of the Bayes-optimal predictor:
\begin{align}
    \text{Div}_x(t) := \frac{1}{2}D_{KL}[p^*(x_t\mid x_{<t})\;\Vert\; &p_\phi(x_t\mid x_{<t})] \\
    + \frac{1}{2}D_{KL}[ p_\phi(x_t\mid x_{<t}) \;&\Vert\; p^*(x_t\mid x_{<t})]\nonumber
\end{align}
\looseness=-1
providing a principled measure of the model's deviation from the ideal predictor at each position in the sequence.
\subsubsection*{The KL bump}
\looseness=-1
We find that the model initially perfectly fits the Bayesian oracle, followed by a characteristic bump at short context, followed by a steady decrease towards an asymptote. That the transient bump was also noticed by 
\citep[Fig. 7]{xie2021explanation}. We hypothesize that this behavior arises because, in very short contexts, the model memorize marginal token statistics, as in \citep{kobayashi2023transformer}. This strategy, however, fails after a few tokens given the exponential growth of possible sequences as a function of length. Aside from this subtlety, we highlight that Transformers trained with next-token loss struggle in regions of high ambiguity.

\subsubsection*{Effect of increased model size}
\looseness=-1
Increasing model size leads to rapid convergence in performance at large sequence lengths, but the KL bump persists across all model scales. This finding underscores a key limitation of current autoregressive models: while they allocate uniform compute per token—often increasing slightly with position—the difficulty of prediction varies across the sequence. In high-ambiguity regions (early in the sequence), the model is under-parameterized relative to the task due to the difficulty of latent posterior inference, while in low-ambiguity regions (later positions), it is over-parameterized. We also attempted to mitigate this imbalance by training on a skewed distribution emphasizing shorter sequences (\cref{fig:skewed}), hoping to increase model focus on high-ambiguity regimes. However, this yielded no significant improvements, suggesting the need for more sophisticated approaches.

\section{Monte-Carlo predictor}
\label{sec:mc}
\looseness=-1
Our core idea is to approximate the Bayesian integral in \cref{eq:oracle} using a Monte Carlo \citep[MC;][]{robert1999monte} estimate: we draw multiple samples from the task posterior  $p(\theta\mid x_{<t})$, compute the conditional predictions $p(x_t\mid x_{<t},\theta)$ for each, and average the results. This approach not only offers a principled mechanism for separately allocating modeling capacity to task inference and next-token prediction during training, but also introduces a natural form of test-time scaling via the number of samples $S$:
\vspace{-3mm}
\begin{align}\label{eq:mc}
    p_{\phi,\psi}(x_t\mid x_{<t}) &= \frac{1}{S}\sum_{i=1}^S p_\phi(x_t\mid x_{<t},\theta_i)\\\text{ where }\theta_i &\overset{\psi}{\sim} \; p(\theta\mid x_{<t})\nonumber
\end{align}\par
\vspace{-3mm}
\looseness=-1
The main challenge lies in implementing this formulation outside synthetic environments (which is still the aim of this paper), where the true latent variable $\theta$ is unknown—making it unclear how to train the components of the predictor. We address this with a three-step solution \cref{fig:mc_pred}\textbf{a)}:
\vspace{-2mm}
\begin{enumerate}
    \vspace{-0.15cm}\item \textbf{Latent proxy} : Replace the (in practice) unknown latent $\theta$ with a contextual embedding  $z=E(x_{ctx})$, where $x_{ctx}\sim p^*_\theta$. Crucially, we restrict $x_{ctx}$ to regimes in which the posterior inference problem would be unambiguous (e.g., large sequence lengths in MetaHMM), such that the learned $z$ can serve as a good proxy for the true task latent. In practice, we use average-pooled hidden states from the frozen pre-trained model $p_{\phi_0}(x_t\mid x_{<t})$ to define $E$.
    \vspace{-0.15cm}\item \textbf{Conditional predictor} : Train a conditional sequence model $p_\phi(x_t\mid x_{<t},z)$ by fine-tuning a pre-trained model $p_{\phi_0}(x_t\mid x_{<t})$, prepending the embedding $z=E(x_{ctx})$ to its input. Sequences $x_{ctx}$ and $x_{1:T}$ are importantly drawn from the same generator $p^*_\theta$.
    \vspace{-0.15cm}\item \textbf{Inference sampler} : Train a diffusion model parameterized by $\psi$ to sample contextual embedding $z\sim p(z\mid x_{<t})$ conditional on $x_{<t}$ for $t\in [1,T]$.
\end{enumerate}
\vspace{-2mm}
\looseness=-1
The key intuition is that the sequence model $p_\phi$ is \textit{only} used for unambiguous prediction (given $z$), while the diffusion model is \textit{only} used for task inference (without access to the token to be predicted). This structure introduces inductive biases tailored to the generative process, while retaining the flexibility of learned amortized inference. Additionally, it is straightforward to derive from standard pre-trained language models through targeted fine-tuning. The idea of training a diffusion model on sentence embeddings is inspired from \cite{lovelace2023latent,lovelace2024diffusion}.
\vspace{-2mm}
\subsection{Results}
\looseness=-1
We perform the aforementioned procedure on the models in \cref{fig:metahmm} with an additional context $x_{ctx}$ of length $100$, ensuring unambiguity of $\theta$. Next we train a Diffusion Transformer \citep[DiT;][]{peebles2023scalable} to sample $z=E(x_{ctx})$ from $x_{<t}$ for $t\in [0,50]$. Note that we deliberately used a larger diffusion than necessary to ensure that capacity of the task inference machinery was not the bottleneck. Studying the total parameter efficiency of our method (i.e. including the diffusion module) is left for future work.
\vspace{-2mm}
\subsubsection*{Improved performance for small models}
\looseness=-1
For small model sizes (e.g. 1.1M), we observe a clear improvement of the Monte-Carlo predictor over $p_{\phi_0}(x_t\mid x_{<t})$,  see \cref{fig:mc_pred}\textbf{b)}. Further, as expected, divergence to oracle monotonically decreases with additional MC samples, demonstrating the test-time scaling potential of our approach. We also confirm the importance of having long sequences $x_{ctx}$ for our latent proxy $z$ in \cref{fig:short_ctx}. Observe that the $\text{Div}$ of the MC predictor for this 1.1M model with $S=5$ is similar to the $\text{Div}$ of the 6.3M model in \cref{fig:metahmm}\textbf{c)}.
\vspace{-2mm}
\subsubsection*{Diminishing returns with scale}
\looseness=-1
However, as model size increases, our method has diminishing returns \cref{fig:mc_pred}\textbf{c)}: for the biggest model, the MC predictor underperforms the traditional model for all number of samples. We attribute this discrepancy to multiple possible causes. 
\looseness=-1
On one hand, as the model size increases, its divergence to the Bayesian oracle at short context decreases, which sets the bar higher for the MC estimate. This reflects the insight from The Bitter Lesson \citep{sutton2019bitter}: as model capacity increases, architectural priors matter less. Our approach is thus most applicable in settings where the underlying sequence model underfits the Bayesian oracle—likely the case for foundation models faced with the complexities of natural language. In such scenarios, our method offers a mechanism for enhancing performance in high-ambiguity regimes without increasing compute for easier parts of the sequence. At the same time, it is also plausible that engineering issues are at play, which we discuss in \cref{sec:details}.
\vspace{-4mm}
\section{Discussion}
\label{sec:related}
\looseness=-1
Using our synthetic benchmark, MetaHMM, we have demonstrated that Transformers fail to approximate the Bayesian posterior predictive in high-ambiguity regimes. As noted, a similar issue is faced by humans, who address it with adaptive behavior \citep{gigerenzer1996reasoning}. This highlights a key limitation of current approaches to sequence modeling, where fixed compute budget is used regardless of contextual uncertainty. We argue that foundation models should be \textit{ambiguity-sensitive}, adapting their inference effort to the difficulty of the prediction.

\looseness=-1
As a step in this direction, we proposed a modular method that bootstraps a standard autoregressive model into a two-stage predictor: a diffusion-based context sampler and a conditional transformer. This architecture enables test-time scalable approximate Bayesian inference through Monte Carlo sampling. Our experiments show improved performance under ambiguity, though further work is needed to improve efficiency and decide if it can be applied to larger models.

\looseness=-1
More broadly, our contribution is to clearly identify a structural problem—handling epistemic uncertainty—and provide a foundation for future solutions. Beyond scalable inference, alternative directions include learned heuristics tailored to ambiguous contexts, and mechanisms for information-seeking behavior. While our framework does not support explicit actions, recent trends in RL-finetuned models \citep{openai2024openaio1card, guo2025deepseek} may be implicitly addressing ambiguity through learned clarification or retrieval behaviors. We hope that this work stimulates a comprehensive integration of past and current research related to the identified ambiguity problem; towards foundation models which go beyond Bayes-optimality \citep{grau2022beyond}.

\bibliography{ref}

\begin{thebibliography}{60}
\providecommand{\natexlab}[1]{#1}
\providecommand{\url}[1]{\texttt{#1}}
\expandafter\ifx\csname urlstyle\endcsname\relax
  \providecommand{\doi}[1]{doi: #1}\else
  \providecommand{\doi}{doi: \begingroup \urlstyle{rm}\Url}\fi

\bibitem[Aky{\"u}rek et~al.(2022)Aky{\"u}rek, Schuurmans, Andreas, Ma, and Zhou]{akyurek2022learning}
Aky{\"u}rek, E., Schuurmans, D., Andreas, J., Ma, T., and Zhou, D.
\newblock What learning algorithm is in-context learning? investigations with linear models.
\newblock \emph{arXiv preprint arXiv:2211.15661}, 2022.

\bibitem[Aky{\"u}rek et~al.(2024)Aky{\"u}rek, Wang, Kim, and Andreas]{akyurek2024context}
Aky{\"u}rek, E., Wang, B., Kim, Y., and Andreas, J.
\newblock In-context language learning: Architectures and algorithms.
\newblock \emph{arXiv preprint arXiv:2401.12973}, 2024.

\bibitem[Binz et~al.(2022)Binz, Gershman, Schulz, and Endres]{binz2022heuristics}
Binz, M., Gershman, S.~J., Schulz, E., and Endres, D.
\newblock Heuristics from bounded meta-learned inference.
\newblock \emph{Psychological review}, 129\penalty0 (5):\penalty0 1042, 2022.

\bibitem[Bommasani et~al.(2021)Bommasani, Hudson, Adeli, Altman, Arora, von Arx, Bernstein, Bohg, Bosselut, Brunskill, et~al.]{bommasani2021opportunities}
Bommasani, R., Hudson, D.~A., Adeli, E., Altman, R., Arora, S., von Arx, S., Bernstein, M.~S., Bohg, J., Bosselut, A., Brunskill, E., et~al.
\newblock On the opportunities and risks of foundation models.
\newblock \emph{arXiv preprint arXiv:2108.07258}, 2021.

\bibitem[Bradbury et~al.(2018)Bradbury, Frostig, Hawkins, Johnson, Leary, Maclaurin, Necula, Paszke, VanderPlas, Wanderman-Milne, et~al.]{bradbury2018jax}
Bradbury, J., Frostig, R., Hawkins, P., Johnson, M.~J., Leary, C., Maclaurin, D., Necula, G., Paszke, A., VanderPlas, J., Wanderman-Milne, S., et~al.
\newblock Jax: composable transformations of python+ numpy programs.
\newblock 2018.

\bibitem[Brown et~al.(2020)Brown, Mann, Ryder, Subbiah, Kaplan, Dhariwal, Neelakantan, Shyam, Sastry, Askell, et~al.]{brown2020language}
Brown, T., Mann, B., Ryder, N., Subbiah, M., Kaplan, J.~D., Dhariwal, P., Neelakantan, A., Shyam, P., Sastry, G., Askell, A., et~al.
\newblock Language models are few-shot learners.
\newblock \emph{Advances in neural information processing systems}, 33:\penalty0 1877--1901, 2020.

\bibitem[Chan et~al.(2022)Chan, Santoro, Lampinen, Wang, Singh, Richemond, McClelland, and Hill]{chan2022data}
Chan, S., Santoro, A., Lampinen, A., Wang, J., Singh, A., Richemond, P., McClelland, J., and Hill, F.
\newblock Data distributional properties drive emergent in-context learning in transformers.
\newblock \emph{Advances in neural information processing systems}, 35:\penalty0 18878--18891, 2022.

\bibitem[Chen et~al.(2022)Chen, Zhang, and Hinton]{chen2022analog}
Chen, T., Zhang, R., and Hinton, G.
\newblock Analog bits: Generating discrete data using diffusion models with self-conditioning.
\newblock \emph{arXiv preprint arXiv:2208.04202}, 2022.

\bibitem[Friston et~al.(2017)Friston, FitzGerald, Rigoli, Schwartenbeck, and Pezzulo]{friston2017active}
Friston, K., FitzGerald, T., Rigoli, F., Schwartenbeck, P., and Pezzulo, G.
\newblock Active inference: a process theory.
\newblock \emph{Neural computation}, 29\penalty0 (1):\penalty0 1--49, 2017.

\bibitem[Gigerenzer \& Goldstein(1996)Gigerenzer and Goldstein]{gigerenzer1996reasoning}
Gigerenzer, G. and Goldstein, D.~G.
\newblock Reasoning the fast and frugal way: models of bounded rationality.
\newblock \emph{Psychological review}, 103\penalty0 (4):\penalty0 650, 1996.

\bibitem[Gilboa et~al.(2009)Gilboa, Postlewaite, and Schmeidler]{gilboa2009always}
Gilboa, I., Postlewaite, A., and Schmeidler, D.
\newblock Is it always rational to satisfy savage's axioms?
\newblock \emph{Economics \& Philosophy}, 25\penalty0 (3):\penalty0 285--296, 2009.

\bibitem[Gottlieb \& Oudeyer(2018)Gottlieb and Oudeyer]{gottlieb2018towards}
Gottlieb, J. and Oudeyer, P.-Y.
\newblock Towards a neuroscience of active sampling and curiosity.
\newblock \emph{Nature Reviews Neuroscience}, 19\penalty0 (12):\penalty0 758--770, 2018.

\bibitem[Grau-Moya et~al.(2022)Grau-Moya, Del{\'e}tang, Kunesch, Genewein, Catt, Li, Ruoss, Cundy, Veness, Wang, et~al.]{grau2022beyond}
Grau-Moya, J., Del{\'e}tang, G., Kunesch, M., Genewein, T., Catt, E., Li, K., Ruoss, A., Cundy, C., Veness, J., Wang, J., et~al.
\newblock Beyond bayes-optimality: meta-learning what you know you don't know.
\newblock \emph{arXiv preprint arXiv:2209.15618}, 2022.

\bibitem[Grau-Moya et~al.(2024)Grau-Moya, Genewein, Hutter, Orseau, Del{\'e}tang, Catt, Ruoss, Wenliang, Mattern, Aitchison, et~al.]{grau2024learning}
Grau-Moya, J., Genewein, T., Hutter, M., Orseau, L., Del{\'e}tang, G., Catt, E., Ruoss, A., Wenliang, L.~K., Mattern, C., Aitchison, M., et~al.
\newblock Learning universal predictors.
\newblock \emph{arXiv preprint arXiv:2401.14953}, 2024.

\bibitem[Graves(2016)]{graves2016adaptive}
Graves, A.
\newblock Adaptive computation time for recurrent neural networks.
\newblock \emph{arXiv preprint arXiv:1603.08983}, 2016.

\bibitem[Guo et~al.(2025)Guo, Yang, Zhang, Song, Zhang, Xu, Zhu, Ma, Wang, Bi, et~al.]{guo2025deepseek}
Guo, D., Yang, D., Zhang, H., Song, J., Zhang, R., Xu, R., Zhu, Q., Ma, S., Wang, P., Bi, X., et~al.
\newblock Deepseek-r1: Incentivizing reasoning capability in llms via reinforcement learning.
\newblock \emph{arXiv preprint arXiv:2501.12948}, 2025.

\bibitem[Hahn \& Goyal(2023)Hahn and Goyal]{hahn2023theoryemergentincontextlearning}
Hahn, M. and Goyal, N.
\newblock A theory of emergent in-context learning as implicit structure induction, 2023.
\newblock URL \url{https://arxiv.org/abs/2303.07971}.

\bibitem[Hendel et~al.(2023)Hendel, Geva, and Globerson]{hendel2023context}
Hendel, R., Geva, M., and Globerson, A.
\newblock In-context learning creates task vectors.
\newblock \emph{arXiv preprint arXiv:2310.15916}, 2023.

\bibitem[Ho \& Salimans(2022)Ho and Salimans]{ho2022classifier}
Ho, J. and Salimans, T.
\newblock Classifier-free diffusion guidance.
\newblock \emph{arXiv preprint arXiv:2207.12598}, 2022.

\bibitem[Izacard \& Grave(2020)Izacard and Grave]{izacard2020distilling}
Izacard, G. and Grave, E.
\newblock Distilling knowledge from reader to retriever for question answering.
\newblock \emph{arXiv preprint arXiv:2012.04584}, 2020.

\bibitem[Keluskar et~al.(2024)Keluskar, Bhattacharjee, and Liu]{keluskar2024llms}
Keluskar, A., Bhattacharjee, A., and Liu, H.
\newblock Do llms understand ambiguity in text? a case study in open-world question answering.
\newblock In \emph{2024 IEEE International Conference on Big Data (BigData)}, pp.\  7485--7490. IEEE, 2024.

\bibitem[Kobayashi et~al.(2023)Kobayashi, Kuribayashi, Yokoi, and Inui]{kobayashi2023transformer}
Kobayashi, G., Kuribayashi, T., Yokoi, S., and Inui, K.
\newblock Transformer language models handle word frequency in prediction head.
\newblock \emph{arXiv preprint arXiv:2305.18294}, 2023.

\bibitem[Kojima et~al.(2022)Kojima, Gu, Reid, Matsuo, and Iwasawa]{kojima2022large}
Kojima, T., Gu, S.~S., Reid, M., Matsuo, Y., and Iwasawa, Y.
\newblock Large language models are zero-shot reasoners.
\newblock \emph{Advances in neural information processing systems}, 35:\penalty0 22199--22213, 2022.

\bibitem[Lampinen et~al.(2024)Lampinen, Chan, Singh, and Shanahan]{lampinen2024broader}
Lampinen, A.~K., Chan, S.~C., Singh, A.~K., and Shanahan, M.
\newblock The broader spectrum of in-context learning.
\newblock \emph{arXiv preprint arXiv:2412.03782}, 2024.

\bibitem[Lewis et~al.(2020)Lewis, Perez, Piktus, Petroni, Karpukhin, Goyal, K{\"u}ttler, Lewis, Yih, Rockt{\"a}schel, et~al.]{lewis2020retrieval}
Lewis, P., Perez, E., Piktus, A., Petroni, F., Karpukhin, V., Goyal, N., K{\"u}ttler, H., Lewis, M., Yih, W.-t., Rockt{\"a}schel, T., et~al.
\newblock Retrieval-augmented generation for knowledge-intensive nlp tasks.
\newblock \emph{Advances in neural information processing systems}, 33:\penalty0 9459--9474, 2020.

\bibitem[Li et~al.(2024)Li, Zhang, Zhang, Zhu, Yu, Zhao, and Heng]{li2024towards}
Li, L., Zhang, H., Zhang, X., Zhu, S., Yu, Y., Zhao, J., and Heng, P.-A.
\newblock Towards an information theoretic framework of context-based offline meta-reinforcement learning.
\newblock \emph{arXiv preprint arXiv:2402.02429}, 2024.

\bibitem[Lieder \& Griffiths(2020)Lieder and Griffiths]{lieder2020resource}
Lieder, F. and Griffiths, T.~L.
\newblock Resource-rational analysis: Understanding human cognition as the optimal use of limited computational resources.
\newblock \emph{Behavioral and brain sciences}, 43:\penalty0 e1, 2020.

\bibitem[Linderman et~al.(2025)Linderman, Chang, Harper-Donnelly, Kara, Li, Duran-Martin, and Murphy]{linderman2025dynamax}
Linderman, S.~W., Chang, P., Harper-Donnelly, G., Kara, A., Li, X., Duran-Martin, G., and Murphy, K.
\newblock Dynamax: A python package for probabilistic state space modeling with jax.
\newblock \emph{Journal of Open Source Software}, 10\penalty0 (108):\penalty0 7069, 2025.

\bibitem[Liu et~al.(2023)Liu, Wu, Michael, Suhr, West, Koller, Swayamdipta, Smith, and Choi]{liu2023we}
Liu, A., Wu, Z., Michael, J., Suhr, A., West, P., Koller, A., Swayamdipta, S., Smith, N.~A., and Choi, Y.
\newblock We're afraid language models aren't modeling ambiguity.
\newblock \emph{arXiv preprint arXiv:2304.14399}, 2023.

\bibitem[Lovelace et~al.(2023)Lovelace, Kishore, Wan, Shekhtman, and Weinberger]{lovelace2023latent}
Lovelace, J., Kishore, V., Wan, C., Shekhtman, E., and Weinberger, K.~Q.
\newblock Latent diffusion for language generation.
\newblock \emph{Advances in Neural Information Processing Systems}, 36:\penalty0 56998--57025, 2023.

\bibitem[Lovelace et~al.(2024)Lovelace, Kishore, Chen, and Weinberger]{lovelace2024diffusion}
Lovelace, J., Kishore, V., Chen, Y., and Weinberger, K.~Q.
\newblock Diffusion guided language modeling.
\newblock \emph{arXiv preprint arXiv:2408.04220}, 2024.

\bibitem[Millidge et~al.(2021)Millidge, Seth, and Buckley]{millidge2021understanding}
Millidge, B., Seth, A., and Buckley, C.
\newblock Understanding the origin of information-seeking exploration in probabilistic objectives for control.
\newblock \emph{arXiv preprint arXiv:2103.06859}, 2021.

\bibitem[Mittal et~al.(2024)Mittal, Elmoznino, Gagnon, Bhardwaj, Sridhar, and Lajoie]{mittal2024does}
Mittal, S., Elmoznino, E., Gagnon, L., Bhardwaj, S., Sridhar, D., and Lajoie, G.
\newblock Does learning the right latent variables necessarily improve in-context learning?
\newblock \emph{arXiv preprint arXiv:2405.19162}, 2024.

\bibitem[Nakano et~al.(2021)Nakano, Hilton, Balaji, Wu, Ouyang, Kim, Hesse, Jain, Kosaraju, Saunders, et~al.]{nakano2021webgpt}
Nakano, R., Hilton, J., Balaji, S., Wu, J., Ouyang, L., Kim, C., Hesse, C., Jain, S., Kosaraju, V., Saunders, W., et~al.
\newblock Webgpt: Browser-assisted question-answering with human feedback.
\newblock \emph{arXiv preprint arXiv:2112.09332}, 2021.

\bibitem[Niwa \& Iso(2024)Niwa and Iso]{niwa2024ambignlg}
Niwa, A. and Iso, H.
\newblock Ambignlg: Addressing task ambiguity in instruction for nlg.
\newblock \emph{arXiv preprint arXiv:2402.17717}, 2024.

\bibitem[Oord et~al.(2018)Oord, Li, and Vinyals]{oord2018representation}
Oord, A. v.~d., Li, Y., and Vinyals, O.
\newblock Representation learning with contrastive predictive coding.
\newblock \emph{arXiv preprint arXiv:1807.03748}, 2018.

\bibitem[OpenAI(2024)]{openai2024openaio1card}
OpenAI.
\newblock Openai o1 system card, 2024.
\newblock URL \url{https://arxiv.org/abs/2412.16720}.

\bibitem[Ortega et~al.(2019)Ortega, Wang, Rowland, Genewein, Kurth-Nelson, Pascanu, Heess, Veness, Pritzel, Sprechmann, et~al.]{ortega2019meta}
Ortega, P.~A., Wang, J.~X., Rowland, M., Genewein, T., Kurth-Nelson, Z., Pascanu, R., Heess, N., Veness, J., Pritzel, A., Sprechmann, P., et~al.
\newblock Meta-learning of sequential strategies.
\newblock \emph{arXiv preprint arXiv:1905.03030}, 2019.

\bibitem[Oudeyer \& Kaplan(2007)Oudeyer and Kaplan]{oudeyer2007intrinsic}
Oudeyer, P.-Y. and Kaplan, F.
\newblock What is intrinsic motivation? a typology of computational approaches.
\newblock \emph{Frontiers in neurorobotics}, 1:\penalty0 108, 2007.

\bibitem[Panwar et~al.(2024)Panwar, Ahuja, and Goyal]{panwar2024incontextlearningbayesianprism}
Panwar, M., Ahuja, K., and Goyal, N.
\newblock In-context learning through the bayesian prism, 2024.
\newblock URL \url{https://arxiv.org/abs/2306.04891}.

\bibitem[Peebles \& Xie(2023)Peebles and Xie]{peebles2023scalable}
Peebles, W. and Xie, S.
\newblock Scalable diffusion models with transformers.
\newblock In \emph{Proceedings of the IEEE/CVF international conference on computer vision}, pp.\  4195--4205, 2023.

\bibitem[Rabiner \& Juang(1986)Rabiner and Juang]{rabiner1986introduction}
Rabiner, L. and Juang, B.
\newblock An introduction to hidden markov models.
\newblock \emph{ieee assp magazine}, 3\penalty0 (1):\penalty0 4--16, 1986.

\bibitem[Rizvi-Martel et~al.(2024)Rizvi-Martel, Lizaire, Lacroce, and Rabusseau]{rizvi2024simulating}
Rizvi-Martel, M., Lizaire, M., Lacroce, C., and Rabusseau, G.
\newblock Simulating weighted automata over sequences and trees with transformers.
\newblock In \emph{International Conference on Artificial Intelligence and Statistics}, pp.\  2368--2376. PMLR, 2024.

\bibitem[Robert et~al.(1999)Robert, Casella, and Casella]{robert1999monte}
Robert, C.~P., Casella, G., and Casella, G.
\newblock \emph{Monte Carlo statistical methods}, volume~2.
\newblock Springer, 1999.

\bibitem[Salimans \& Ho(2022)Salimans and Ho]{salimans2022progressive}
Salimans, T. and Ho, J.
\newblock Progressive distillation for fast sampling of diffusion models.
\newblock \emph{arXiv preprint arXiv:2202.00512}, 2022.

\bibitem[Sanborn et~al.(2010)Sanborn, Griffiths, and Navarro]{sanborn2010rational}
Sanborn, A.~N., Griffiths, T.~L., and Navarro, D.~J.
\newblock Rational approximations to rational models: Alternative algorithms for category learning.
\newblock \emph{Psychological Review}, 117\penalty0 (4):\penalty0 1144--1167, 2010.
\newblock \doi{10.1037/a0020511}.

\bibitem[Schick et~al.(2023)Schick, Dwivedi-Yu, Dess{\`\i}, Raileanu, Lomeli, Hambro, Zettlemoyer, Cancedda, and Scialom]{schick2023toolformer}
Schick, T., Dwivedi-Yu, J., Dess{\`\i}, R., Raileanu, R., Lomeli, M., Hambro, E., Zettlemoyer, L., Cancedda, N., and Scialom, T.
\newblock Toolformer: Language models can teach themselves to use tools.
\newblock \emph{Advances in Neural Information Processing Systems}, 36:\penalty0 68539--68551, 2023.

\bibitem[Snell et~al.(2024)Snell, Lee, Xu, and Kumar]{snell2024scaling}
Snell, C., Lee, J., Xu, K., and Kumar, A.
\newblock Scaling llm test-time compute optimally can be more effective than scaling model parameters.
\newblock \emph{arXiv preprint arXiv:2408.03314}, 2024.

\bibitem[Sutton(2019)]{sutton2019bitter}
Sutton, R.
\newblock The bitter lesson.
\newblock \emph{Incomplete Ideas (blog)}, 13\penalty0 (1):\penalty0 38, 2019.

\bibitem[Todd et~al.(2023)Todd, Li, Sharma, Mueller, Wallace, and Bau]{todd2023function}
Todd, E., Li, M.~L., Sharma, A.~S., Mueller, A., Wallace, B.~C., and Bau, D.
\newblock Function vectors in large language models.
\newblock \emph{arXiv preprint arXiv:2310.15213}, 2023.

\bibitem[Vaswani et~al.(2017)Vaswani, Shazeer, Parmar, Uszkoreit, Jones, Gomez, Kaiser, and Polosukhin]{vaswani2017attention}
Vaswani, A., Shazeer, N., Parmar, N., Uszkoreit, J., Jones, L., Gomez, A.~N., Kaiser, {\L}., and Polosukhin, I.
\newblock Attention is all you need.
\newblock \emph{Advances in neural information processing systems}, 30, 2017.

\bibitem[Von~Oswald et~al.(2023)Von~Oswald, Niklasson, Randazzo, Sacramento, Mordvintsev, Zhmoginov, and Vladymyrov]{von2023transformers}
Von~Oswald, J., Niklasson, E., Randazzo, E., Sacramento, J., Mordvintsev, A., Zhmoginov, A., and Vladymyrov, M.
\newblock Transformers learn in-context by gradient descent.
\newblock In \emph{International Conference on Machine Learning}, pp.\  35151--35174. PMLR, 2023.

\bibitem[Wang et~al.(2023)Wang, Zhu, Saxon, Steyvers, and Wang]{wang2023large}
Wang, X., Zhu, W., Saxon, M., Steyvers, M., and Wang, W.~Y.
\newblock Large language models are latent variable models: Explaining and finding good demonstrations for in-context learning.
\newblock In \emph{Thirty-seventh Conference on Neural Information Processing Systems}, 2023.
\newblock URL \url{https://openreview.net/forum?id=BGvkwZEGt7}.

\bibitem[Wei et~al.(2022)Wei, Wang, Schuurmans, Bosma, Xia, Chi, Le, Zhou, et~al.]{wei2022chain}
Wei, J., Wang, X., Schuurmans, D., Bosma, M., Xia, F., Chi, E., Le, Q.~V., Zhou, D., et~al.
\newblock Chain-of-thought prompting elicits reasoning in large language models.
\newblock \emph{Advances in neural information processing systems}, 35:\penalty0 24824--24837, 2022.

\bibitem[Xie et~al.(2021)Xie, Raghunathan, Liang, and Ma]{xie2021explanation}
Xie, S.~M., Raghunathan, A., Liang, P., and Ma, T.
\newblock An explanation of in-context learning as implicit bayesian inference.
\newblock \emph{arXiv preprint arXiv:2111.02080}, 2021.

\bibitem[Yang et~al.(2025)Yang, Lin, Lee, Papailiopoulos, and Nowak]{yang2025task}
Yang, L., Lin, Z., Lee, K., Papailiopoulos, D., and Nowak, R.
\newblock Task vectors in in-context learning: Emergence, formation, and benefit.
\newblock \emph{arXiv preprint arXiv:2501.09240}, 2025.

\bibitem[Yao et~al.(2023)Yao, Zhao, Yu, Du, Shafran, Narasimhan, and Cao]{yao2023react}
Yao, S., Zhao, J., Yu, D., Du, N., Shafran, I., Narasimhan, K., and Cao, Y.
\newblock React: Synergizing reasoning and acting in language models.
\newblock In \emph{International Conference on Learning Representations (ICLR)}, 2023.

\bibitem[Zhang \& Choi(2023)Zhang and Choi]{zhang2023clarify}
Zhang, M.~J. and Choi, E.
\newblock Clarify when necessary: Resolving ambiguity through interaction with lms.
\newblock \emph{arXiv preprint arXiv:2311.09469}, 2023.

\bibitem[Zhu(2021)]{zhu2021leebert}
Zhu, W.
\newblock Leebert: Learned early exit for bert with cross-level optimization.
\newblock In \emph{Proceedings of the 59th Annual Meeting of the Association for Computational Linguistics and the 11th International Joint Conference on Natural Language Processing (Volume 1: Long Papers)}, pp.\  2968--2980, 2021.

\bibitem[Zhuang et~al.(2024)Zhuang, Singh, Liu, Shang, and Gao]{zhuang2024vector}
Zhuang, Y., Singh, C., Liu, L., Shang, J., and Gao, J.
\newblock Vector-icl: In-context learning with continuous vector representations.
\newblock \emph{arXiv preprint arXiv:2410.05629}, 2024.

\end{thebibliography}
\bibliographystyle{icml2025}

\appendix
\onecolumn
\section{Extented related work}

\subsection{Sequence Meta-Learning and In-Context Learning}

The link between in-context learning (ICL) and meta-learning was established early on \citep{brown2020language} and has become central to understanding foundation model behavior. \citet{xie2021explanation} were among the first to formalize this connection by framing foundation model pre-training as meta-learning over a distribution of tasks. This perspective has since inspired a wide range of studies \citep{von2023transformers, hahn2023theoryemergentincontextlearning, chan2022data, akyurek2022learning, akyurek2024context, grau2024learning}, most of which focus on few-shot ICL \citep{lampinen2024broader}, where the model infers a classification function in-context: $x_0, y_0, \ldots, x_q \rightarrow y_q$.

Our work instead focuses on sequence meta-learning, a more general—and we argue more accurate—framing of what foundation models do : model sequences. In this view, each sequence is generated by a latent task $\theta$, and the model must perform implicit inference over $\theta$ to make accurate predictions.

Several synthetic distributions have been proposed for studying sequence meta-learning, but, as shown in \cref{fig:metahmm}, none satisfy all the properties we require. GINC \citep{xie2021explanation} uses sequences drawn from a mixture of HMMs but lacks compositional structure and evaluates models mainly through predictive accuracy. RegBench \citep{akyurek2024context} also evaluates models on probabilistic sequence data and compares them to a Bayesian oracle, but focuses on architecture comparisons (e.g., RNNs vs. Transformers) rather than the ambiguity-computation mismatch we investigate. Other works study non-Markovian sequence distributions, such as those based on PCFGs \citep{hahn2023theoryemergentincontextlearning} or Turing machines \citep{grau2024learning}, but these lack a tractable Bayesian oracle, limiting their utility for quantitative evaluation.

\subsection{Latent variables in Transformers}
Multiple previous works have explored to what extent Transformers explicitly represent the latent variables underlying an in-context problem. \cite{todd2023function, hendel2023context} have shown that in some few-show ICL tasks, such word associations, Transformer indeed represent the task latent $\theta$, allowing for manipulation of the inference process. \cite{yang2025task} have expanded on the conditions necessary for \textit{task vectors} to appear, and how they often don't. Authors have also proposed a method to force the appearance of task vectors using an auxiliary loss. \cite{mittal2024does} also demonstrated that task vectors sometimes do not appear in function approximation few-shot ICL tasks and attributed it to the fact that Transformers had trouble representing the functional form $p^*_\theta(x\mid y)$. Lastly, \cite{zhuang2024vector} have trained Transformers with continuous task vectors in order to increase the performance on some ICL tasks.

\subsection{Ambiguity and the Limits of Bayesian Inference}
The challenge of inference under ambiguity has long been studied in cognitive science and decision theory. While Bayesian inference offers a normative ideal, exact inference becomes intractable—or even behaviorally irrational—when the posterior is broad \citep{gigerenzer1996reasoning, gilboa2009always, lieder2020resource}.

To account for how humans reason effectively despite such limitations, resource-rational analysis posits that people optimize a trade-off between accuracy and cognitive cost \citep{lieder2020resource}. Rather than marginalizing over all hypotheses, humans rely on heuristics or approximate inference mechanisms—often learned through experience—that deliver fast, "good enough" solutions \citep{sanborn2010rational, binz2022heuristics}.

Another human strategy is information-seeking, where agents act to reduce uncertainty before committing to a belief or decision. This is formalized in active inference \citep{friston2017active}, in which agents take epistemic actions—e.g., querying, exploring, or deferring—to gather evidence and improve predictions. Related ideas appear in reinforcement learning under the banner of curiosity and intrinsic motivation \citep{oudeyer2007intrinsic, gottlieb2018towards}.

These insights motivate our central hypothesis: when the posterior over latent tasks $p(\theta \mid x_{<t})$ is broad, prediction becomes not only statistically harder, but also computationally more demanding. While humans respond flexibly through heuristics and exploration, transformers by default apply a fixed computation budget to all inputs. 

\subsection{Existing solutions}

Although rarely framed in terms of latent task inference, many approaches in the foundation model literature implicitly address ambiguity. One prominent strategy is retrieval-augmented generation (RAG) \citep{lewis2020retrieval, izacard2020distilling}, which enriches the input context with relevant documents, enabling models to disambiguate queries using external knowledge. RAG is especially helpful in settings with underspecified or ambiguous inputs.

Other approaches equip models with clarification-seeking capabilities \citep{zhang2023clarify}, allowing them to request more information before answering. Similarly, tool-augmented models \citep{schick2023toolformer, yao2023react} can interact with APIs or calculators, actively reducing uncertainty—akin to epistemic actions in humans. Further, system prompts, such as those used in chatbots, serve to disambiguate the model's role and task \citep{niwa2024ambignlg}.

As a side note, chain-of-thought prompting \citep{wei2022chain, kojima2022large} methods improve reasoning by encouraging intermediate steps. However, theses methods enhance can be seen as addressing cases where the conditional prediction $p(x_t \mid x_{<t}, \theta)$, not the task inference, is too difficult for the model.  

Finally, reinforcement learning from human feedback (RLHF) can lead to emergent ambiguity-sensitive behaviors \citep{openai2024openaio1card, guo2025deepseek}. Models fine-tuned with RLHF sometimes defer responses, ask clarifying questions, or invoke tools—particularly when direct answers are penalized for inaccuracy \citep{nakano2021webgpt}. These behaviors align with active inference \citep{friston2017active, millidge2021understanding}, though ambiguity resolution is not explicitly optimized.

\subsection{Test-Time Scaling and Adaptive Inference}

Our method also contributes to the growing literature on test-time scalable inference, where computation is dynamically adjusted based on input difficulty. Early work on adaptive computation time (ACT) \citep{graves2016adaptive} allowed models to learn how many steps to take. In transformers, this has evolved into early exiting mechanisms \citep{zhu2021leebert}, which conditionally terminate processing. See \citet{snell2024scaling} for a modern discussion of test-time scaling. Our approach offers a specific, and principled interpretation of test-time scaling from a Bayesian perspective.

\section{Additional details}
\label{sec:details}

\subsection{Engineering challenges of the Monte-Carlo predictor}
As model size increases, the latent representation $z$ also grows in dimensionality, potentially making the diffusion sampling task more difficult. Despite the DiT’s large capacity, it may still underfit. Improved diffusion design—e.g., via classifier-free guidance \citep{ho2022classifier} or self-conditioning \citep{chen2022analog}—could help bridge this gap. Additionally, the embedding $z$ may encode high- frequency details from $x_{ctx}$ that are hard to sample accurately. This issue could be exacerbated by increasing the dimension of $x$. A promising direction would be to regularize $z$ to capture only information relevant to task identity $\theta$. For example, maximizing the mutual information $I(z;\theta)$ via a contrastive loss \citep{oord2018representation, li2024towards} could encourage more robust low-dimensional. We leave a full exploration of these directions to future work, viewing our method as a first step toward separating task inference and prediction in foundation models.

\looseness=-1
\subsection{MetaHMM details}
All experiments in this paper are performed on three seeds of a MetaHMM with a fixed size. HMMs have a hidden state space of dimension $20$ and an observational space of size $50$. Other hyperparamters are described in \cref{table:hparams}a). The total number of HMMs can be computed as
\begin{align}
    \underbrace{(n_b\cdot s_b \cdot d_b)}_{\text{Base cycles}}\cdot\underbrace{(g_f^{n_f}\cdot d_f \cdot s_f)}_{\text{Cycle families}}\cdot\underbrace{(\alpha_e ^ {g_e} \cdot \beta_e)}_{\text{Emission groups}}
\end{align}
which in our case gives 12,288 HMMs.

\subsection{Architecture details}
All causal transformers use the Adam optimizer with learning rate $0.001$, batch size $256$ and $50,000$ updates. Further hyperparameters of the causal Transformers are given by \cref{table:hparams}b). When training on a MetaHMM environment, we hold out 1000 HMMs ($\sim \frac{1}{12}$ of all) for validation, and report validation metrics throughout.

All diffusion models are DiTs \citep{peebles2023scalable} trained with the Adam optimizer with learning rate $0.0001$, batch size $512$ and $100,000$ updates. The conditioning information, i.e. $x_{<t}$, is first passed through an Transformer encoder (without causal masking) and then both through a cross-attention block and a adaLN-Zero block. Hyperparameters are given by \cref{table:hparams}c). When training, we hold out $\frac{1}{10}$th of all HMMs for validation. The DiT uses the velocity parameterization \citep{salimans2022progressive} with an $L2$ loss and a cosine noise schedule. Sampling is performed using the DDPM sampler with 50 timesteps.  Other hyperparameters are the same as in \citep{lovelace2023latent}.

\begin{figure}
\hfill
\subfigure[\textbf{MetaHMM}]{\begin{tabular}{|ll|}
\hline
\textbf{Base cycles}                            &   \\ \hline
\multicolumn{1}{|l|}{Cycles $n_b$}             & 4 \\ \hline
\multicolumn{1}{|l|}{Step-size $s_b$}          & 2 \\ \hline
\multicolumn{1}{|l|}{Directions $d_b$}         & 2 \\ \hline
\textbf{Cycle families}                           &   \\ \hline
\multicolumn{1}{|l|}{Families $n_f$}           & 3 \\ \hline
\multicolumn{1}{|l|}{Groups per family $g_f$}  & 2 \\ \hline
\multicolumn{1}{|l|}{Directions $d_f$}         & 2 \\ \hline
\multicolumn{1}{|l|}{Step-sizes $s_f$}         & 2 \\ \hline
\textbf{Emission matrix}                       &   \\ \hline
\multicolumn{1}{|l|}{Groups $g_e$}             & 3 \\ \hline
\multicolumn{1}{|l|}{Emission per group $\alpha_e$} & 2 \\ \hline
\multicolumn{1}{|l|}{Shifts $\beta_e$}             & 3 \\ \hline
\end{tabular}}
\hfill
\subfigure[\textbf{Causal Transformer}]{\begin{tabular}{|l|c|c|c|}
\hline
Size/Hyperparmeters & \multicolumn{1}{l|}{Layers} & \multicolumn{1}{l|}{Heads} & \multicolumn{1}{l|}{Dimension} \\ \hline
1.1M                & 4                           & 4                          & 128                            \\ \hline
2.0M                & 6                           & 6                          & 128                            \\ \hline
6.3M                & 6                           & 8                          & 256                            \\ \hline
25.0M               & 8                           & 8                          & 512                            \\ \hline
\end{tabular}}
\hfill
\subfigure[\textbf{DiT}]{\begin{tabular}{|c|c|}
\hline
DiT Layers  & 8   \\ \hline
DiT Heads   & 8   \\ \hline
Dimension   & 512 \\ \hline
Enc. Layers & 8   \\ \hline
Enc. Heads  & 8   \\ \hline
\end{tabular}}
\hfill
\caption{\textbf{Hyperparameters}}
\label{table:hparams}
\end{figure}

\subsection{Figures}

\begin{figure}[!ht]
  \centering
    \includegraphics[width=0.9\linewidth]{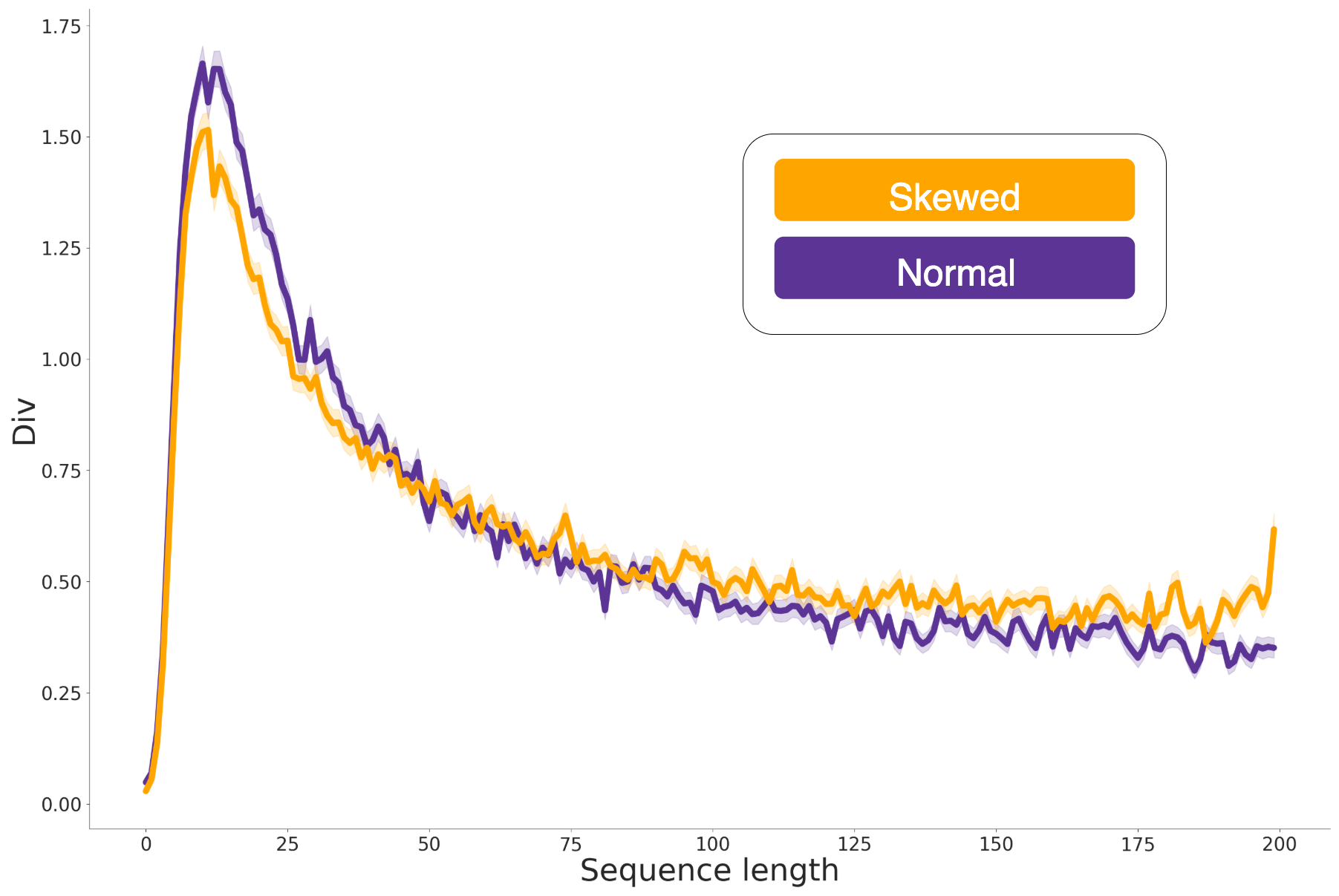} 
    \caption{\textbf{Transformer trained on random sequence lengths} Variation of the next-token training setup of \cref{fig:metahmm}c) where sequences have random lengths uniformly sampled between 1 and 200. This puts more pressure on the predictors to perform well at low context lengths. Batch size are adjusted so that the amount of tokens per batch remains constant (and equal to full-length training).}
    \label{fig:skewed}
\end{figure}

\begin{figure}[!ht]
  \centering
    \includegraphics[width=0.9\linewidth]{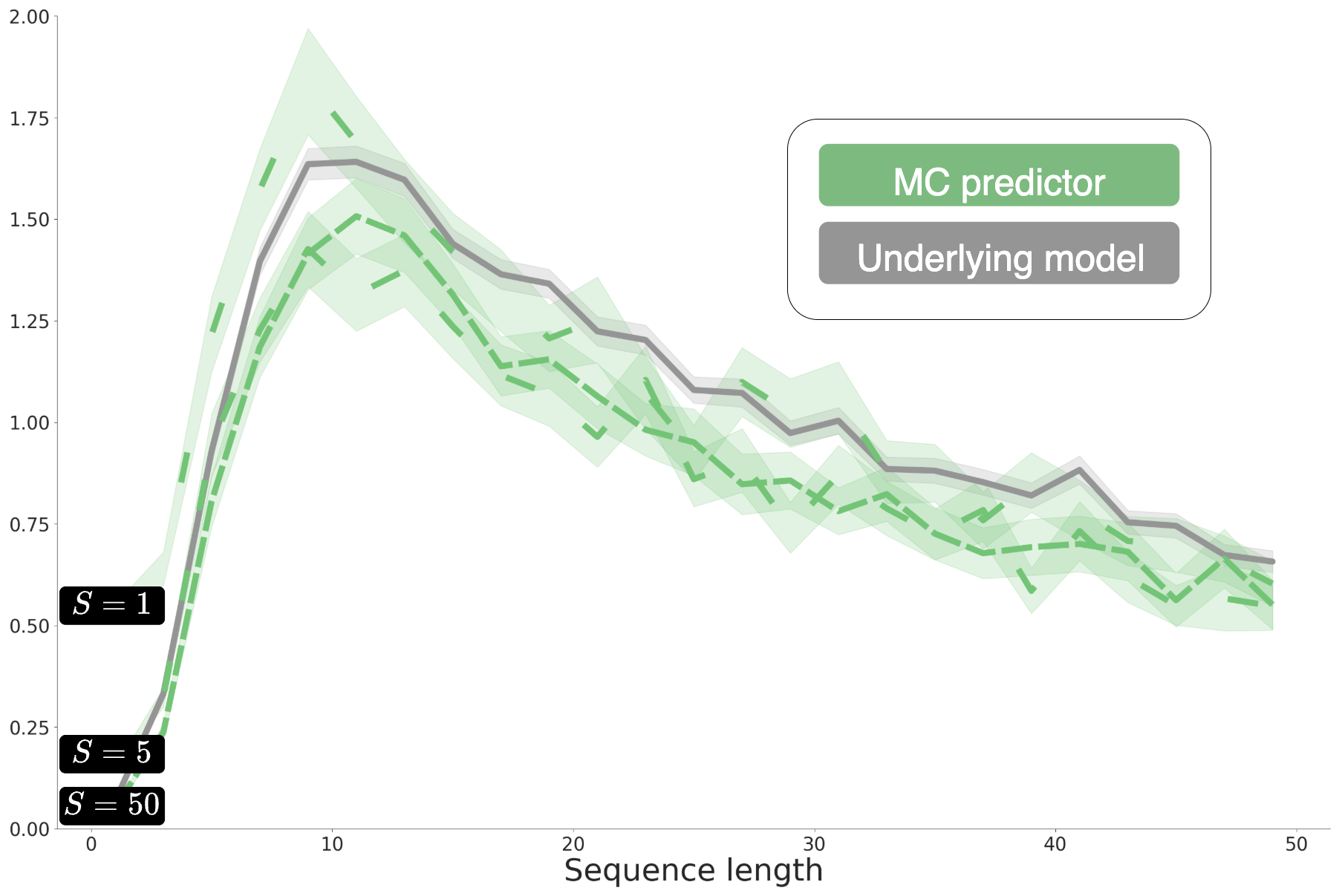} 
    \caption{\textbf{Short $x_{ctx}$} variation of \cref{fig:mc_pred}b) where $x_{ctx}$ has length 10. This means that $z$ should be a poor proxy for $\theta$ and the MC predictor should do poorly.} 
    \label{fig:short_ctx}
    \vspace{-5mm}
\end{figure}
\end{document}